\documentclass[
]{ceurart}

\usepackage{listings}
\usepackage{times}
\usepackage{color,soul}
\usepackage{latexsym}
\usepackage{epigraph}
\begin{document}

\copyrightyear{2024}
\copyrightclause{Copyright for this paper by its authors.
  Use permitted under Creative Commons License Attribution 4.0
  International (CC BY 4.0).}

\conference{CREAI 2024 – Workshop on Artificial Intelligence and Creativity, October 19–24, 2024, Santiago de Compostela, Spain}
\title{A Perspective on Literary Metaphor in the Context of Generative AI}

\author[1,2]{Imke {van Heerden}}[%
orcid=0000-0002-4224-8800,
email=imke.vanheerden@kcl.ac.uk
]
\address[1]{Department of Digital Humanities, King’s College London, UK}
\address[2]{Department of Comparative Literature, Koç University, Turkey}

\author[3,4]{Anil Bas}[%
orcid=0000-0002-3833-6023,
email=abas@bournemouth.ac.uk
]
\address[3]{National Centre for Computer Animation, Bournemouth University, UK}
\address[4]{Department of Computer Engineering, Faculty of Technology, Marmara University, Turkey}

\cortext[1]{Work partially done while the authors were at their previous institutions.}
\cortext[2]{An earlier version of this work was presented at an NAACL workshop (non-archival)}

\begin{abstract}
At the intersection of creative text generation and literary theory, this study explores the role of literary metaphor and its capacity to generate a range of meanings. In this regard, literary metaphor is vital to the development of any particular language. To investigate whether the inclusion of original figurative language improves textual quality, we trained an LSTM-based language model in Afrikaans. The network produces phrases containing compellingly novel figures of speech. Specifically, the emphasis falls on how AI might be utilised as a defamiliarisation technique, which disrupts expected uses of language to augment poetic expression. Providing a literary perspective on text generation, the paper raises thought-provoking questions on aesthetic value, interpretation and evaluation.
\end{abstract}

\begin{keywords}
  Computational Creativity \sep
  Creative Text Generation \sep
  Afrikaans \sep
  Metaphor \sep
  Figurative Language
\end{keywords}

\maketitle

\epigraph{\hfill --- \footnotesize{\textit{lewe in hierdie nuwe hande waar ek algoritmies kuier}} \\ \footnotesize{[\textit{life in these new hands where I socialise algorithmically}]}}

\section{Introduction}

Although creative text generation has made notable strides in recent years \cite{lau2018deep,zugarini2019neural,shihadeh2020emily,van2020automatic,chakrabarty2021mermaid,kobis2021artificial,uthus2021augmenting,van2021afriki,yang2021fudge, bas2022a}, the question of literary value requires further consideration. Large Language Models such as ChatGPT \cite{openai2022chatgpt} have become a lens for public discussions on generative AI. The topic is understandably controversial in writing and publishing communities, given deep concern over IP, copyright and privacy violations \cite{edwards2024private}. Are all AI systems intrinsically problematic, or could emerging technologies be designed in more principled and useful ways? What constitutes a responsible approach to Natural Language Generation (NLG) in creative contexts?

Another important question motivating this research is whether AI can enrich language. Literary metaphor is one avenue by which to address the subject, as a formative component of linguistic creativity and a renowned resource for the generation of novel meaning \cite{hidalgo2019metaphor}. Drawing on metaphor theory, this study suggests that the inclusion of figurative forms enhances the perception of originality, emotivity and memorability as well as facilitates connection. The aim is by no means to deceive through human likeness, but to engage with the exciting potential of literary metaphors to introduce new ways of using language \cite{semino2008metaphorlit}. To explore this idea, we propose an LSTM-based network for figurative language generation in Afrikaans. The proposed model produces intriguing phrases with distinctive figures of speech, such as metaphor, simile and personification.

Metaphor is integral to everyday language \cite{lakoff1980conceptual}. However, this work does not pursue the creation of conventional metaphor. Instead, we prioritise creative metaphor with artistic merit, as typically expressed in poetic discourse \cite{caracciolo2016creative}. Compared to metaphors in everyday speech, unusual and unexpected figures stand out \cite{kovecses2010metaphor} and, in a literary context, capture readers’ attention \cite{steen1994understanding}. Conducting an example evaluation of our generated output, using prevalent evaluation criteria, we argue that current frameworks are unable to do justice to poetic language. This is directly connected to the evaluation problem in creative text generation \cite{hamalainen2021human}. Our problematisation of NLG evaluation frameworks encourages task-specific evaluation criteria.

Finally, an important contribution lies in the choice of language. Although Multilingual Large Language Models such as GPT-4 \cite{openai2024gpt} and Llama 3 \cite{meta2024llama} are able to generate text in the Afrikaans language, studies \cite{van2003improving,sanby2016comparing,ziering2016towards,dirix2017universal} and datasets \cite{eiselen2014developing,augustinus2016afribooms,roux2016south} that directly focus on Afrikaans are limited. This resonates with other low-resource languages as well. In \cite{nekoto2020participatory,orife2020masakhane,adelani2021masakhaner}, Masakhane shows that Natural Language Processing (NLP) research in African languages is under-represented. NLP systems are currently dominated by a handful of languages \cite{joshi2020state}, and Afrikaans is one of many across the world presently unable to match their progress and sophistication. On the upside, we believe that low-resource languages offer exciting opportunities for experimentation, collaboration and growth. The sustained invention of new metaphors is a clear indication that a language is alive \cite{trask2004language}.

\section{Literary Metaphor}

Defined as ``language that is more expressive and/or poetic than referential in its linguistic function'' \cite{chandler2011dictionary}, figurative language comprises metaphor, metonymy, simile, personification and various other figures of speech. Metaphor can be defined as “the phenomenon whereby we talk and, potentially, think about something in terms of something else” \cite{semino2008metaphor}, revealing connections between concepts that are not necessarily apparent. It is a vital resource of creative writing \cite{baldick1996concise} often associated with originality. In fact, original metaphor is considered ``the controlling element in all creative language'' \cite{newmark1988textbook}. 

Literary metaphor is challenging to delineate given its relation to literary theory, which is vast and complex \cite{freeman2011role}. Though vital to acknowledge, it is beyond this paper’s scope to sufficiently account for the range and depth of contemporary metaphor theory. For the purposes of this discussion, our emphasis falls on inventive uses of language whilst recognising the inherent difficulty in objectively differentiating literary from ordinary language. The (dis)continuity between literary and nonliterary metaphor is subject to much debate, but there is some scholarly consensus that the former tends to exhibit more creativity, novelty, originality, complexity and interpretive difficulty \cite{semino2008metaphorlit}. Foregrounding its imaginative nature, Gibbs aptly describes metaphor as the ``dreamwork of language'' \cite{gibbs1990process}.

\section{Related Work}

Studies in figurative text generation include simile \cite{harmon2015figure8,chakrabarty2020generating}, slogan \cite{alnajjar2018talent} and metaphor \cite{gero2019metaphoria,brooks2020discriminative,chakrabarty2021mermaid} generation. These computational approaches involve style transfer and word masking but also, more traditionally, non-computational theories of metaphor creation, e.g. the tenor-vehicle model \cite{richards1936philosophy}. It must be noted that related work on figurative language tends to focus on English and other resource-rich languages, using knowledge bases, graphs, pretrained networks and datasets.

Our study differs in three primary ways. First, we train an LSTM-based model from scratch, without any restrictions, constraints or classifiers. We do not borrow any large-scale language models \cite{devlin2019bert,lewis2020bart} for fine-tuning, or large-scale corpora \cite{williams2018broad} for textual entailment. Regardless, these options are not available to low-resource languages. In this sense, our network's advantage lies in its simplicity. Second, we approach the creative text generation process from a literary perspective, investigating ways in which literature and NLG might inform and benefit from one another. Third, we discuss the challenge of evaluating figurative language and emphasise the importance of well-defined task-specific criteria.

\section{Approach}

We use a two-layer vanilla LSTM architecture \cite{hochreiter1997long}, which consists of two LSTM layers with dropout layers, a fully connected layer and a softmax layer. The model was trained on a single literary novel titled \textit{Die biblioteek aan die einde van die wêreld} (literally, \textit{The Library at the End of the World}) \cite{van2019biblioteek}. The English translation is published as \textit{A Library to Flee} \cite{van2022library}. Regarding the text's suitability as dataset, the book abounds in vivid imagery and figurative expressions, and includes Afrikaans varieties as well as some English. Broadly, this follows the same approach as \cite{van2021afriki}. In this paper, however, we explicitly focus and expand on fully automatic text generation, centring figurative language in particular. Moreover, we investigate the vital role of original metaphor in creative writing. To clarify the process, the generation of text is open-ended (referred to as unconditional text generation); user-provided inputs, known as text prompts, are not provided. In the model, we do not enforce any specific rules, model constraints, components or use complex training schemes.

In AI research, technological innovation is understandably prized. In a creative context, however, this work serves to challenge the assumption that technical state-of-the-art equates to aesthetic value. In other words, scientific excellence does not necessarily correlate with artistic excellence. The following section describes metaphor's creative potential to inspire new modes of expression.

\begin{table*}
  \caption{Example results of figurative language generation. The translations might not do justice to the original given the distinctiveness of some of the formulations as well as the compounding nature of Afrikaans.}
  \label{table:examples} 
  \begin{tabular}{l|l}
\toprule
\textbf{Original (Afrikaans)} & \textbf{Translation (English)}\\
\midrule

ons biblioteek by die werkwoord gekaap                  & our library hijacked at the verb\\
die wêreld sê ek met boeke                              & the world I say with books\\
wêreldletterkunde in armoede                            & world literature in poverty\\
saggies soos ’n spokerigheid                            & softly like a ghostliness \\
in die vlug van papier                                  & in the flight of paper\\
sy vingers draai om haar gevoel                         & his fingers wrap around her feeling\\
ek het ’n gloeiende noordgrens                          & I have a glowing northern border\\
woede is jou mond                                       & anger is your mouth\\
brand my in die oggendlug                               & burn me in the morning air\\
die wind stoppelbaard vorentoe                          & the wind stubbles forth\\
sy kyk verras op, sy oë verlate                         & she looks up in surprise, his eyes deserted\\
verandering speel as foto’s van die wind                & change plays as photos of the wind\\
die petrolbomme wat nie vertel nie                      & the petrol bombs that do not tell\\
sukkel is hulle kuns                                    & struggling is their art\\
onbeskermde skittering in die woord                     & unprotected brightness in the word\\
my rug se wit greep                                     & my back’s white grip\\
ek is geld want niks kan bloei nie                      & I am money because nothing can bleed\\
aarselend weerskante van die staar                      & hesitant on either side of the stare\\
demokrasie was ’n daktuin                               & democracy was a roof garden\\
\small{begin die sonsopkoms voor die dak van my gesig}  & \small{begins the sunrise before the roof of my face}\\
\small{jou uitgespoel is ’n onderstebo losgewoel} & \small{your rinsed-out is an upside-down tossed-loose}\\
\small{gesprekke vir die oomblik skoongeskraap bleek}   & \small{conversations momentarily clean-scraped pale}\\
\bottomrule
\end{tabular}
\end{table*}

\section{Figurative Language Generation in Afrikaans}

Table~\ref{table:examples} provides examples of generated phrases and sentences containing figures of speech such as metaphor, simile and personification. (Note that punctuation and capitalisation were removed in some instances.) Similar to the trained data, the network outputs unique descriptive formulations.

Given our interest in creative text generation, we believe that the success of the results is not determined by the amount of similarities shared between referents, as explained by Giles et al.~\cite{giles1991metaphor}. Instead, we adopt Black's interaction theory of metaphor \cite{black1962models}. This is relevant to the study given its emphasis on the generative function of figurative language. To clarify, because literary works (and, by implication, literary metaphors) are open to interpretation, they are ``capable of generating a whole range of possible meanings. They do not so much contain meaning as produce it'' \cite{eagleton2014read}. In this view, meaning is not static; metaphor does not draw on pre-existing likeness but instead \emph{creates} new, often surprising, likeness between concepts \cite{indurkhya1992metaphor,veale2016metaphor}.

Consequently, metaphorical language invites the reader to participate in the process of meaning-making \cite{white1996structure}, thus facilitating connection between reader and text \cite{cohen1978metaphor}. As regards connection, Veale~\cite{veale2019metaphor} emphasises metaphor's interactive dimension: it draws an engaged response from the listener/reader. Furthermore, Gibbs et al. suggest that original metaphors ``communicate more emotional intensity than conventional metaphor'' \cite{gibbs2002s}. It follows that literary metaphor is related to not only heightened creativity but emotion as well
\cite{fainsilber1987metaphorical,lubart1997emotion,fussell1998figurative}. As a result, we prioritise unexpected associations between disparate concepts, e.g. ``democracy'' as a ``roof garden'' (see Table~\ref{table:examples}, line 19). Through this metaphor, dissimilar domains \emph{interact}, giving rise to unpredictable connections and perspectives \cite{way1991knowledge}. It is challenging to measure originality in generated text \cite{klebanov2020automated}. However, if figurative meaning does involve a ``mismatch'' between domains \cite{fogelin2011figuratively}, one might argue: the greater the mismatch, the greater the novelty. The effect is that of defamiliarisation, a formalist technique (referring to Russian formalism, a school of literary theory) that “makes language strange”, inviting readers to see the habitual world in fresh new ways \cite{shklovsky1917art}. Using AI in this sense does not replicate patterns but rather challenges convention.

\section{Remarks on Evaluation}

Evaluation is a well-known problem in NLG that, we believe, intensifies in creative text generation. Howcroft et al.~\cite{howcroft2020twenty} draw attention to the wide and varied range of evaluation criteria in NLG, and the necessity of standardisation – a sentiment echoed by Celikyilmaz et al.~\cite{celikyilmaz2020evaluation}. As a potential counterargument, Hämäläinen and Alnajjar state~\cite{hamalainen2021great}: if a study’s problem definition, method and evaluation are not aligned, its evaluation results inevitably lack value. Addressing the issue from a literary perspective, Van Heerden and Bas~\cite{van2021ai} propose the reconceptualisation of evaluation methods for creative systems, in particular. They argue that, though suited to standard text generation, commonly used frameworks do not encapsulate the nuances of poetic language and form.

In our study, similarly, there is a clear misalignment between our objectives and frequently used evaluation categories. To illustrate this, we apply to our results a selection of prevalent categories identified by Van der Lee et al.~\cite{van2020human} – specifically \textit{fluency}, \textit{coherence}, \textit{accuracy} and \textit{informativeness}. To clarify, although these are intended for text generation in general, we chose them since there are no standardised evaluation frameworks for creative text. Applying the criteria, the generated output appears grammatically satisfied. However, it often lacks internal consistency and accuracy – as in, for instance, “I am money because nothing can bleed” (see Table~\ref{table:examples}, line 17). Referring to the same example, the sentence cannot strictly be regarded as informative (though it does communicate meaning). Although applicability certainly depends on the definition and explanation of the criteria (which are frequently absent), it is likely that this set would bear negative results in evaluation. Nevertheless, the generated text arguably draws surprising and intriguing connections between disparate domains, which we find promising in a poetic context. Analysing the theatrical production \textit{AI: When a Robot Writes a Play}, Van Heerden, Duman and Bas \cite{van2023performing} argue that the script’s distinctive qualities may be positively attributed to the model’s limitations, as opposed to its seamless performance.

Hämäläinen and Alnajjar~\cite{hamalainen2021human} identify commonly evaluated features in creative text generation, including \textit{meaning}, \textit{syntactic correctness}, \textit{novelty}, \textit{relevance} and \textit{emotional value}. These criteria could be more suitable, but we agree that further research, i.e. ``evaluation of evaluation'' \cite{hamalainen2021human}, is first required, specifically to determine whether an evaluation framework could speak to literary concerns and measure a text's potential to captivate readers. Creative text generation holds interdisciplinary appeal, and the standardisation of NLG frameworks does not necessarily preclude the development of other approaches and applications.

\section{Conclusion}

In response to Boden’s pivotal question ``what aesthetically interesting results can computers generate, and how?'' \cite{boden2012creativity}, this study presents an LSTM-based model for figurative language generation in Afrikaans. In general terms, our research demonstrates that a boutique language model can enhance discursive creativity. The paper discusses the importance of metaphor to creative writing, presenting one possible means of achieving enhanced emotivity, depth and originality in generated text. Moreover, figurative language generation is used as a point of departure to reflect on possible intersections between NLG and literary frames of reference. For example, unlike some approaches in the field, we do not expect our output to meet predetermined criteria. Instead, we shift emphasis to the competence of the evaluation framework itself. Questions that may be posed in this regard include: What kind of engagement does this kind of text invite? Does the text possess any compelling qualities, and to what extent are current evaluation criteria able to value these qualities? How could these aspects guide the reconceptualisation of evaluation in creative text generation?

This study’s confluence of technical and poetic practices served as the foundation for South Africa’s, and possibly Africa's, first AI poetry collection, \textit{Silwerwit in die soontoe: Afrikaans se eerste KI‐gedigte} [\textit{Silverwhite into the Distance: Afrikaans' First AI Poetry}] \cite{van2023silwerwit}. Described as “a watershed moment in the Afrikaans poetry tradition” \cite{bennett2023die}, the book explores the creative possibilities of language with the assistance of AI. This experimental work tests the limits of the literary, in the tradition of electronic literature \cite{hayles2016electronic}. The human poet interwove phrases of generated text to create verse, probing how generative AI might augment and challenge the art of poetry. The ambition was to create an original work of literature, primarily, as well as contribute to responsible NLG research.

In an attempt to establish best practice, the book's creators acquired a writer's permission to use his manuscript as training data. For this reason, the front cover credits the writer alongside the developer and the poet. The introduction of this poetry collection explains the distinct roles of the human poet and the AI model in the creative process. It is noteworthy that the first work of AI-generated literature in the region is an instance of co-creativity that privileges the human author. This approach poses an ethical, albeit modest, alternative to LLMs in creative contexts.

\begin{acknowledgments}
Results incorporated in this paper have received funding from the European Union's Horizon 2020 research and innovation programme under the Marie Skłodowska-Curie grant agreement No 900025. This work was also supported by the 2232 International Fellowship for Outstanding Researchers Program of TÜBİTAK [project number 18C285]; and the Engineering and Physical Sciences Research Council [grant number EP/Y009800/1].

The authors are grateful to Etienne van Heerden for sharing the original manuscript of his novel.
\end{acknowledgments}

\bibliography{ref}

\end{document}